\pdfoutput=1

\documentclass[11pt]{article}

\usepackage[final]{acl}

\usepackage{times}
\usepackage{latexsym}

\usepackage[T1]{fontenc}

\usepackage[utf8]{inputenc}

\usepackage{microtype}

\usepackage{inconsolata}

\usepackage{xcolor}
\usepackage{graphicx}
\usepackage[utf8]{inputenc}
\usepackage{CJKutf8}
\usepackage[flushleft]{threeparttable}
\usepackage{amsmath}
\usepackage{booktabs}
\usepackage{amsfonts}
\usepackage{multirow}
\usepackage{inconsolata}
\usepackage{booktabs}
\usepackage{pifont}
\newcommand{\dataset}{KwaiChat}
\newcommand{\cmark}{\ding{51}}%
\newcommand{\xmark}{\ding{55}}%

\title{\dataset: A Large-Scale Video-Driven Multilingual Mixed-Type \\Dialogue Corpus}

\author{
  Xiaoming Shi\textsuperscript{1}, 
  Zeming Liu\textsuperscript{2}\thanks{\quad Corresponding author.}, 
  Yiming Lei\textsuperscript{2}\thanks{Work done during the internship at Kuaishou Technology}, 
  Chenkai Zhang\textsuperscript{2}\footnotemark[2], 
  Haitao Leng\textsuperscript{3,$\ddagger$}, \\ 
  \textbf{Chuan Wang\textsuperscript{4}, 
    Qingjie Liu\textsuperscript{2}\footnotemark[1], 
    Wanxiang Che\textsuperscript{5}, 
    Shaoguo Liu\textsuperscript{3}, 
    Size Li\textsuperscript{3}, 
    Yunhong Wang\textsuperscript{2}}
  \\
  \textsuperscript{1} East China Normal University, Shanghai, China\quad
  \textsuperscript{2} Beihang University, Beijing, China \\
  \textsuperscript{3} KuaiShou Inc., Hangzhou, China\quad
  \textsuperscript{4} Beijing JiaoTong University, Beijing, China \\
  \textsuperscript{5} Research Center for Social Computing and Information Retrieval, HIT, Harbin, China \\
  {\tt xmshi@cs.ecnu.edu.cn}; 
  {\tt \{zmliu, ymlei, ckzhang, qjliu, yhwang\}@buaa.edu.cn} \\
  $\ddagger$ Project leader
}

\begin{document}
\maketitle
\begin{abstract}
Video-based dialogue systems, such as education assistants, have compelling application value, thereby garnering growing interest.
However, the current video-based dialogue systems are limited by their reliance on a single dialogue type, which hinders their versatility in practical applications across a range of scenarios, including question-answering, emotional dialog, etc.
In this paper, we identify this challenge as how to generate video-driven multilingual mixed-type dialogues.
To mitigate this challenge, we propose a novel task and create a human-to-human 
video-driven multilingual mixed-type dialogue corpus, termed {\dataset}, containing a total of 93,209 videos and 246,080 dialogues, across 4 dialogue types, 30 domains, 4 languages, and 13 topics.
Additionally, we establish baseline models on {\dataset}. 
An extensive analysis of 7 distinct LLMs on {\dataset} reveals that GPT-4o achieves the best performance but still cannot perform well in this situation even with the help of in-context learning and fine-tuning, 
which indicates that the task is not trivial and needs further research.\footnote{Our dataset is publicly available at https://github.com/Stan-lei/KwaiChat-NAACL2025.}
\end{abstract}

\section{Introduction}
Recently, groundbreaking advancements in large language models (LLMs)~\cite{zeng2022glm,openai2023gpt4,touvron2023llama,bao2023disc}, have ignited significant technological shifts within the realm of dialogue systems. 
These LLMs are capable of producing responses that closely mimic human conversation, achieving remarkable performance, and fundamentally transforming the underpinnings of dialogue systems. 
Consequently, they are being integrated into various applications, such as healthcare~\cite{laranjo2018conversational,he2023survey,hadi2023survey,li2024healthcare}, education~\cite{lan2024teachers,lieb2024student}, and law~\cite{shu2024lawllm}.
Among these applications, multi-participant video-based dialogue systems stand out for their substantial contributions across various scenarios, such as 
collaborative learning~\cite{Anjewierden2007toward,Trausan2007visual} and entertainment~\cite{herring2009fast,cadilhac2012annotating}, which hold immense practical value and have garnered considerable interest from both academia and industry.

\begin{figure*}[ht]
	\small
	\centering
	\includegraphics[width=0.9\linewidth]{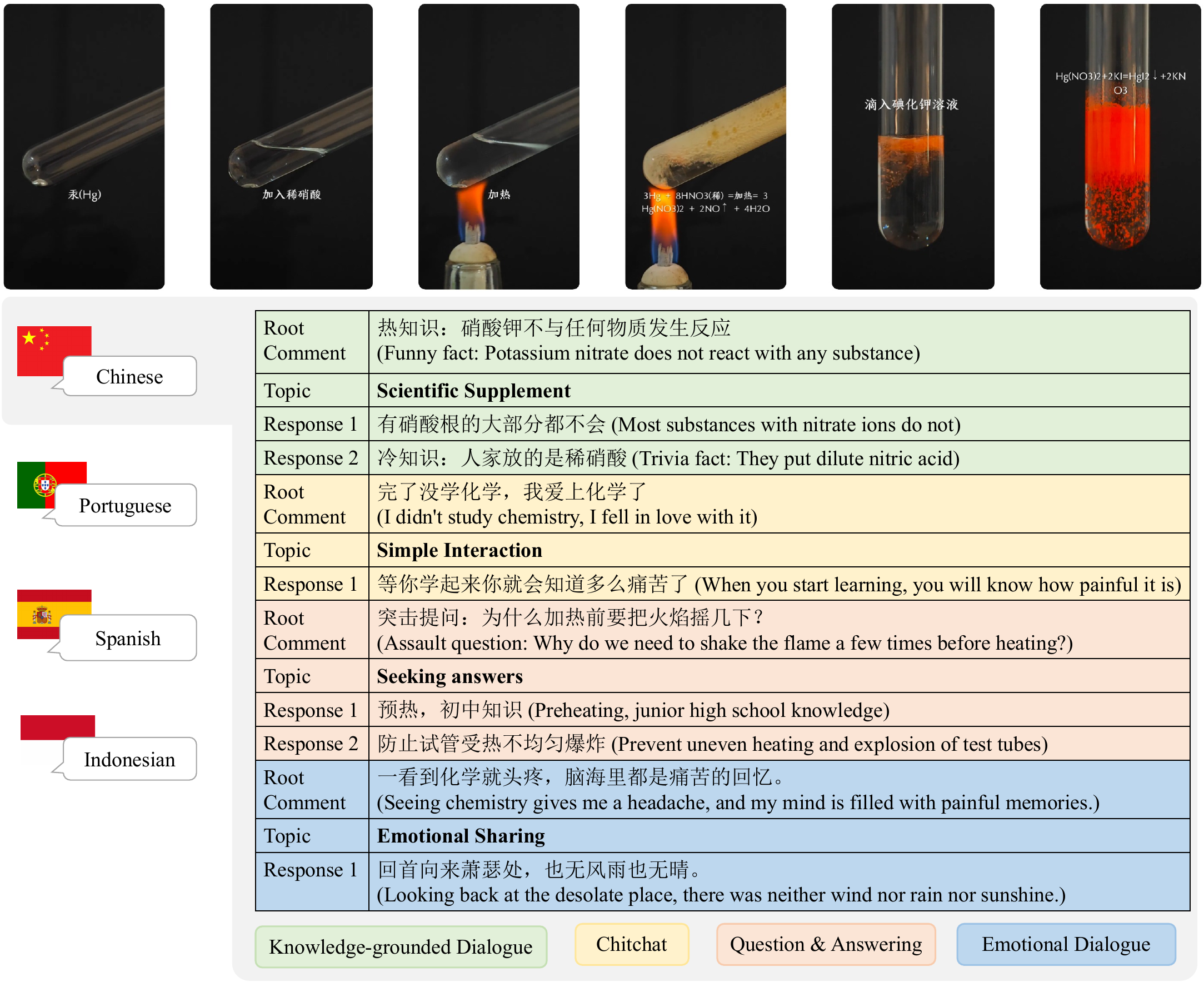}\\
	\caption{An example of {\dataset}. The image above is captured from a video. Below the video, there are comments in four languages, and a Chinese dialogue is shown, with annotated topics and corresponding dialogue types.}
	\label{figure:sample}
\end{figure*}

The current video-based dialogue systems~\cite{lalanne2023livechat} for multiple participants assume that dialogues are performed in a single dialogue type, predominantly focusing on chitchat.
However, in many real-world applications, there are multiple dialogue types in human-bot conversations (termed as mixed-type dialogs), such as chitchat, emotional dialogue, and even question-answering~\cite{ram2018conversational,zhou2020design}.
As illustrated in Figure~\ref{figure:sample} and Figure~\ref{figure:sample_other}, the conversation encompasses a variety of dialogue types, including chitchat, question-answering, knowledge-grounded dialogue, and emotional dialogue. 
Therefore, how to generate video-driven mixed-type multi-participant dialogues is important and non-trivial.
However, the challenge remains unexplored in previous works.

To facilitate the study of video-driven mixed-type multi-participant dialogue generation, 
we first construct a new human-to-human video-driven mixed-type multi-participant dialogue corpus by collecting videos, comments, replies of comments, domains, and topics from a video-sharing platform Kwai\footnote{https://www.kuaishou.com},
owing to the rich video-driven mixed-type multi-participant dialogue resource in Kwai.
For ethical considerations, we initially secure permission to access the data and meticulously remove content that could compromise privacy.
Additionally, in our pursuit to advance multilingual research and enhance the robustness of dialogue systems, we collect a diverse corpus encompassing a variety of languages, including Portuguese, Chinese, Indonesian, and Spanish.
Besides, to ensure the dataset quality and address the long-tail distribution of video topics, 
three filtering strategies and an adaptive video balancing method are utilized, respectively.
Finally, we have constructed a dataset, termed {\dataset}, which encompasses 93,209 videos and 246,080 dialogues. Specifically, KwaiChat spans 4 dialogue types, encompasses 30 different domains, 4 languages, and 13 topics.
As the example shown in Figure~\ref{figure:sample},
multiple participants engage in conversations around a given video in 4 different languages,
and make their responses in different dialogue types based on the video, context, and topics.

To promote research on video-driven mixed-type multi-participant dialogue, we conduct benchmarking experiments on {\dataset} for end-to-end dialogue generation.
Specifically, we conduct an extensive analysis of 7 LLMs' zero-shot (directly utilizing LLMs for inference and showing no case in input), in-context learning (show 3, 4, and 5 cases in input, respectively), fine-tuning performance on {\dataset}.
Experimental results reveal that the existing state-of-the-art LLMs still cannot perform well in this situation even with the help of in-context learning and fine-tuning.

There are three main contributions of this work:
\begin{itemize}
    \item We identify a new challenge as how to generate video-driven multilingual mixed-type multi-participant dialogues.
    \item To promote research on the challenge, we collect a new dataset, termed {\dataset}, containing a total of 93,209 videos and 246,080 dialogues, across 4 dialogue types, 30 domains, 4 languages, and 13 topics.
    \item We conduct extensive experiments of 7 LLMs on {\dataset}. Experiment results show that current LLMs can not perform well in generating video-driven mixed-type multi-participant dialogues even with in-context learning and fine-tuning, which indicates that the task is not trivial and needs further research.
\end{itemize}

\begin{table*}[t]
\centering
\small
\begin{threeparttable}
\begin{tabular}{@{}lccccc@{}}
\toprule
Dataset                & Mixed-type           & Multilingual           & Multimodal               & Multi-parti     & \# of Turns           \\ \midrule
VisDial~\cite{das2017visual}    &  \textcolor{red}{\xmark}   & \textcolor{red}{\xmark}~(EN)        & \textcolor{green}{\cmark}~(image,text)     & \textcolor{red}{\xmark}         & 2.47M              \\
IGC~\cite{mostafazadeh2017image}  &   \textcolor{red}{\xmark}  & \textcolor{red}{\xmark}~(EN)        & \textcolor{green}{\cmark}~(image,text)     & \textcolor{red}{\xmark}         & 25.3K            \\
Image-Chat~\cite{shuster2018image}  &   \textcolor{red}{\xmark} & \textcolor{red}{\xmark}~(EN)        & \textcolor{green}{\cmark}~(image,text)     & \textcolor{red}{\xmark}         & 401K              \\
PhotoChat~\cite{zang2021photochat} &   \textcolor{red}{\xmark}  & \textcolor{red}{\xmark}~(EN)        & \textcolor{green}{\cmark}~(image,text)     & \textcolor{red}{\xmark}         & 156K              \\
MMDD~\cite{lee2021constructing}   &   \textcolor{red}{\xmark} & \textcolor{red}{\xmark}~(EN)        & \textcolor{green}{\cmark}~(image,text)     & \textcolor{red}{\xmark}         & 346K              \\
DialogCC~\cite{lee2022dialogcc}  &   \textcolor{red}{\xmark} & \textcolor{red}{\xmark}~(EN)        & \textcolor{green}{\cmark}~(image,text)     & \textcolor{red}{\xmark}         & 929K               \\
MMChat~\cite{zheng2021mmchat}   &   \textcolor{red}{\xmark}   & \textcolor{red}{\xmark}~(ZH)        & \textcolor{green}{\cmark}~(image,text)     & \textcolor{red}{\xmark}         & 314K              \\
MMDialog~\cite{feng2022mmdialog}  &   \textcolor{red}{\xmark} & \textcolor{red}{\xmark}~(EN)        & \textcolor{green}{\cmark}~(image,text)     & \textcolor{red}{\xmark}         & 4.92M            \\
OpenViDial~\cite{meng2020openvidial} &   \textcolor{red}{\xmark}  & \textcolor{red}{\xmark}~(EN)      & \textcolor{green}{\cmark}~(image,text)     & \textcolor{red}{\xmark}         & 41.10M            \\
OpenViDial 2.0~\cite{wang2021openvidial}  & \textcolor{red}{\xmark}   & \textcolor{red}{\xmark}~(EN)        & \textcolor{green}{\cmark}~(image,text)     & \textcolor{red}{\xmark}         & 5.60M              \\
AVSD~\cite{alamri2019audio}    &  \textcolor{red}{\xmark}    & \textcolor{red}{\xmark}~(EN)        & \textcolor{green}{\cmark}~(video,text)     & \textcolor{red}{\xmark}         & 236K              \\                                 
YTD-18M~\cite{han2023champagne}  &  \textcolor{red}{\xmark}  & \textcolor{red}{\xmark}~(EN)        & \textcolor{green}{\cmark}~(video,text)     & \textcolor{red}{\xmark}         & 54M               \\  
MLED~\cite{poria2018meld}      &  \textcolor{red}{\xmark}    & \textcolor{red}{\xmark}~(EN)        & \textcolor{green}{\cmark}~(video,text)     & \textcolor{red}{\xmark}         & 13.7K               \\  
M$^{3}$ED~\cite{poria2018meld}   &  \textcolor{red}{\xmark}  & \textcolor{red}{\xmark}~(ZH)        & \textcolor{green}{\cmark}~(video,text)     & \textcolor{red}{\xmark}         & 24.4K               \\  
TikTalk~\cite{lin2023tiktalk}   &   \textcolor{red}{\xmark}   & \textcolor{red}{\xmark}~(ZH)        & \textcolor{green}{\cmark}~(video,text)     & \textcolor{red}{\xmark}         & 827K              \\  
       \midrule
CRD3~\cite{rameshkumar2020storytelling} & \textcolor{red}{\xmark} & \textcolor{red}{\xmark}~(EN)     & \textcolor{red}{\xmark}~(text)  & \textcolor{green}{\cmark}    & 399K \\
MediaSum~\cite{zhu2021mediasum}   & \textcolor{red}{\xmark}  & \textcolor{red}{\xmark}~(EN)        & \textcolor{red}{\xmark}~(text)     & \textcolor{green}{\cmark}    & 463K \\
Reddit~\cite{schrading2015analysis} & \textcolor{red}{\xmark} & \textcolor{red}{\xmark}~(EN)        & \textcolor{red}{\xmark}~(text)     & \textcolor{green}{\cmark}      & 628K \\ \midrule
DuRecDial 2.0~\cite{liu-etal-2021-du}  &  \textcolor{green}{\cmark}   & \textcolor{green}{\cmark}~(ZH, EN)     & \textcolor{red}{\xmark}~(text)          & \textcolor{red}{\xmark}   & 255K    \\
COD~\cite{majewska2023cross}         &  \textcolor{red}{\xmark}     & \textcolor{green}{\cmark}~(RU, AR, ID, KIS)        & \textcolor{red}{\xmark}~(text)  & \textcolor{red}{\xmark}    & 800     \\ 
CCNet~\cite{hung2022multi2woz} & \textcolor{red}{\xmark} & \textcolor{green}{\cmark}~(AR, ZH, RU, DE)  & \textcolor{red}{\xmark}~(text)   & \textcolor{red}{\xmark}   & 100K       \\ 
BiToD~\cite{lin2021bitod}  &  \textcolor{red}{\xmark} & \textcolor{green}{\cmark}~(ZH, EN) & \textcolor{red}{\xmark}~(text)      & \textcolor{red}{\xmark}   & 144K     \\ 
MedDialog~\cite{zeng-etal-2020-meddialog}  & \textcolor{red}{\xmark} & \textcolor{green}{\cmark}~(ZH, EN)  & \textcolor{red}{\xmark}~(text)   & \textcolor{red}{\xmark}    & 4M  \\ \midrule
LiveChat~\cite{lalanne2023livechat}   & \textcolor{red}{\xmark} & \textcolor{red}{\xmark}~(JPN)        & \textcolor{green}{\cmark}~(video,text)     & \textcolor{green}{\cmark}         & 3.2M  \\ \midrule
KwaiChat (Our)   &  \textcolor{green}{\cmark} & \textcolor{green}{\cmark}~(ZH, ES, ID, POR)   &  \textcolor{green}{\cmark}~(video, text)  & \textcolor{green}{\cmark}   &     0.8M         \\ \bottomrule
\end{tabular}
\end{threeparttable}
\caption{Comparison of KwaiChat with other dialogue datasets. ``Multi-party'' stands for multi-participant. ``DE'', ``EN'', ``ZH'', ``JPN'', ``ID'', ``RUS'', ``AR'', ``KIS'', ``ES'', and ``POR'' stand for German, English, Chinese, Japanese, Indonesian, Russian, Arabic, Kiswahili, Spanish and Portuguese.}
\label{tab:data_compare}
\end{table*}

\section{Related Work}
This work is closely related to image-based dialogue, video-based dialogue, and mixed-type dialogue.
A summary is shown in Table~\ref{tab:data_compare}.

\subsection{Image-based Dialogue}
Image-based dialogue systems are required to answer questions accurately,
given an image, a dialog context, and a question about the image.
Typical image-based dialogue datasets include VisDial~\cite{das2017visual}, IGC~\cite{mostafazadeh2017image}, Image-Chat~\cite{shuster2018image}, PhotoChat~\cite{zang2021photochat}, MMDD~\cite{lee2021constructing}, DialogCC~\cite{lee2022dialogcc}, MMChat~\cite{zheng2021mmchat}, MMDialog~\cite{feng2022mmdialog}, OpenViDial~\cite{meng2020openvidial}, and OpenViDial 2.0~\cite{wang2021openvidial}.
The dialogues in these datasets are monolingual, image-based, and have two participants.
Compared with these studies, {\dataset} is a unique video-driven multi-participant dialogue corpus that supports multiple languages and incorporates various dialogue types.

\subsection{Video-based Dialogue}
Typical video-based dialogue datasets include AVSD~\cite{alamri2019audio}, YTD-18M~\cite{han2023champagne}, MLED~\cite{poria2018meld}, M$^{3}$ED~\cite{poria2018meld}, TikTalk~\cite{lin2023tiktalk}, LiveChat (Japan)~\cite{lalanne2023livechat}, LiveChat (Chinese)~\cite{gao2023livechat}, and VideoChat~\cite{li2023videochat}.
Dialogues in these datasets are monolingual, video-based, and have two participants.
Compared with the existing literature, {\dataset} stands out as a comprehensive, multi-participant dialogue dataset that boasts multilingual support and diverse dialogue types.


\subsection{Mixed-type Dialogue}
Recently, research on the mixed-type dialogue has increased significantly.
These researches fall into two categories: (1) train an all-in-one conversation model by using multiple single-skill conversation datasets, such as persona-chat, task-oriented dialogue, to bind multiple dialogue skills~\cite{wang2024surveydatasynthesisaugmentation,madotto2020attention,blender2021,Madotto2021TheAA}; (2) collect mixed-type dialog datasets~\cite{Kurt2020,smith-etal-2020,Sun2021AddingCT,liu-etal-2021-du,chiu-etal-2022,liu-etal-2022} to train mixed-type dialog models. Those datasets are intended to mix different dialogue skills to meet specific needs, such as recommending movies and songs, and are unable to solve medical consultations.
Compared with them, we collect a video-driven multilingual mixed-type dialogue corpus, {\dataset}, to facilitate further study.

\section{Dataset Collection}
Kwai\footnote{https://www.kuaishou.com/} is a global video website with a large number of users in various languages and various marked topics, which is suitable for collecting dialogue corpus.
Thus, Kwai is utilized as the data source.
This section describes the two main steps for {\dataset} construction:
(1) dialogue selection;
(2) data merging and balancing.
The dialogue selection is designed to select data of high quality.
Then, to address the long-tail distribution of video topics, data merging, and balancing is applied with an adaptive video balancing method.

\subsection{Dialogue Selection} 


To ensure better data quality, two data filtering strategies are applied, including preliminary filtering based on SQL tags, and heuristics rules.
Besides, to remove data related to violence, illegal activities, and ethical issues, the LLMs-based method is utilized.
Below is a detailed introduction.

\noindent
\textbf{SQL Tags based Filtering}.
The filtering method based on SQL tags is designed to conduct a preliminary screening of data based on SQL tags.
Main comments and reply comments are handled with different strategies.
For main comments, both likes and direct comments must reach a certain threshold, which are set as 500, and 200, respectively.
Besides, those deleted comments, or comments with ``@'', or AI comments are all not kept.
For those reply comments, likes must reach a certain threshold of 50, with those deleted, or with ``@'', or AI comments not kept.
Besides, multiparticipant dialogues are extracted with the requirement that \textit{reply\_to\_comment\_id=root\_comment\_id}.
Finally, a total of 691,282 Chinese dialogues and 623,934 dialogues in other languages are obtained.

\noindent
\textbf{Heuristic Rule-based Filtering}.
Five heuristic rules are utilized for the data filtering.
First, main comments with empty replies are deleted.
If the main comment has no reply comments, delete the main comment.
The rule is designed to obtain multiparticipant, multi-turn dialogues.
Second, short comments are deleted.
If the reply comment has fewer than 3 characters, delete the reply comment.
The rule is designed based on the observation that short replies are mostly tone words with little information content, which is low in quality, such as ``Haha'', and ``like''.
To ensure the dialogue quality, these dialogues are deleted.
Third, comments containing specific keywords are deleted.
Reply comments deleted, which contain the specified keywords [``Feeling will be popular'', ``Will be popular'', ``Occupying the floor'', ``Occupying a floor'', ``Front row'', ``First floor'', ``Second floor''].
Then, data deduplication is applied.
Reply comments with the same content are only kept once.
Finally, a total of 440,666 Chinese dialogues and 303,062 dialogues in other languages are kept.

\noindent
\textbf{LLM based Filtering}.
LLM prompting technique is utilized to filter the dataset.
Specifically, dialogues with the following six aspects are removed by ChatGPT~\cite{chatgpt}. 
1. Illegal Content
\textbf{Violence}: Content describing or inciting violent behavior.
\textbf{Terrorism}: Content related to terrorist activities.
\textbf{Illegal Activities}: Including but not limited to drug trafficking, hacking attacks, copyright infringement, etc.
2. Harmful Content
\textbf{Hate Speech}: Attacks or defamation against groups of race, religion, gender, sexual orientation, nationality, etc.
\textbf{Discrimination}: Any form of discriminatory content.
\textbf{Harassment}: Harassment, threats, or bullying against individuals or groups.
3. Inappropriate Content
\textbf{Adult Content}: Including pornography, obscenity, or explicit sexual content.
\textbf{Extreme Violence}: Content that is too bloody or cruel.
\textbf{Sensitive Topics}: Topics such as suicide, self-harm, or eating disorders.
4. False Information
\textbf{Fake News}: Spreading misleading or false news information.
\textbf{Misleading Information}: Spreading unverified rumors or misleading information.
5. Morality and Ethics
\textbf{Moral Standards}: Content that violates the moral standards recognized by society.
\textbf{Ethical Issues}: Content involving ethical controversies, such as human experimentation or animal abuse.
6. Privacy
\textbf{Personal Information}: Disclosing personal privacy information, such as addresses, phone numbers, ID numbers, etc.
\textbf{Sensitive Data}: Involving sensitive data such as financial information, medical records, etc.

Finally, a total of 391,036 dialogues are kept.

\begin{figure}[t]
	\small
	\centering
	\includegraphics[width=\linewidth]{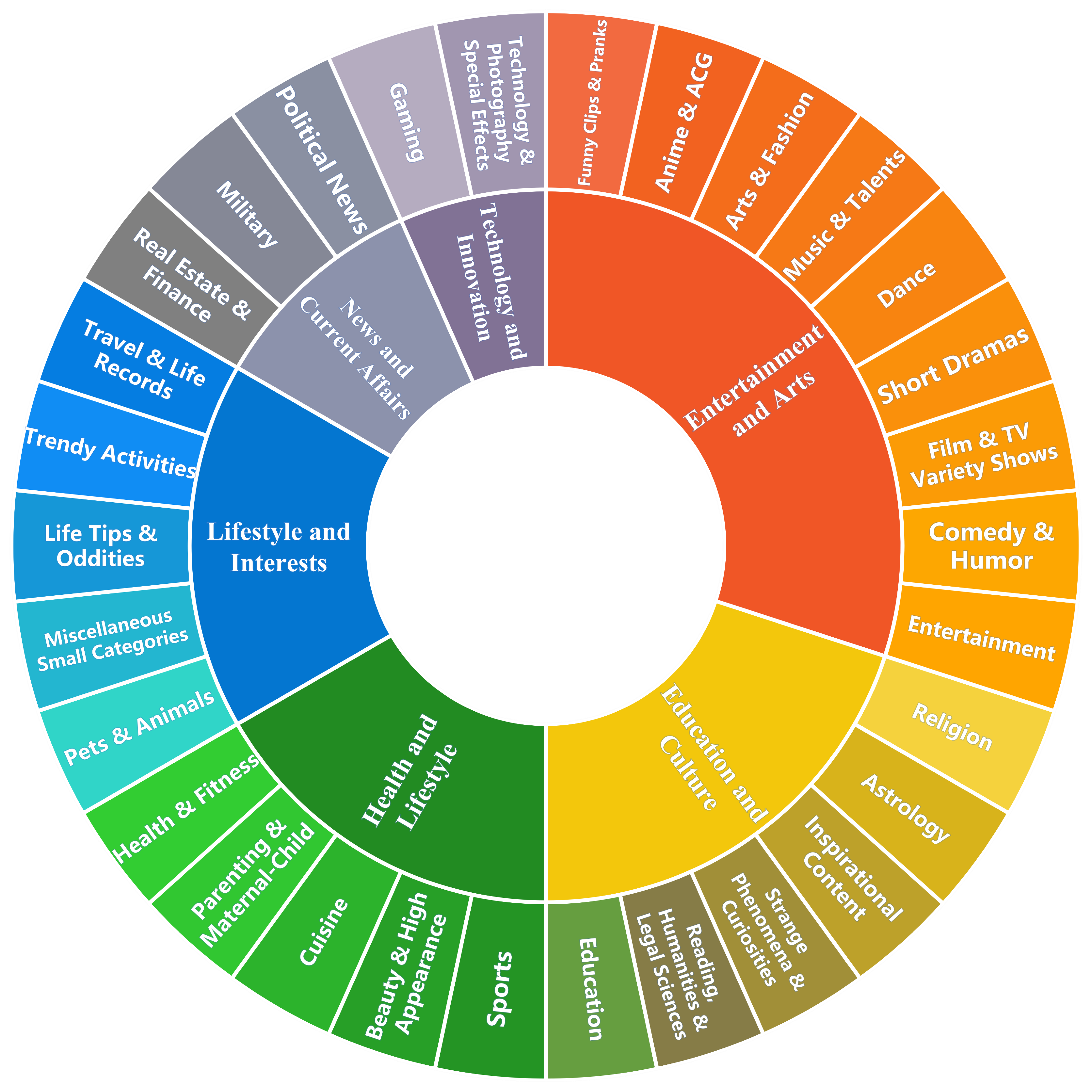}\\
	\caption{Domains of {\dataset}.}
	\label{figure:category}
\end{figure}

\subsection{Data Merging and Balancing}
To address the long-tail distribution of video topics,
an adaptive video topic balancing method based on the number of likes is designed to balance the topics of videos. 
It is observed that an abundance of videos on topics tends to attract a larger audience, which in turn generates a higher volume of views. 
Consequently, this leads to an increase in the number of likes received. 
Given this correlation, it becomes evident that a higher threshold for content curation is necessary to ensure quality and relevance.
Based on this hypothesis, an adaptive algorithm is designed to calculate the like count threshold \textit{Like\_threshold} based on the number of videos in a topic.
The parameters of the algorithm include (1) \textit{base\_threshold} ($\theta_{b}$): The base threshold value, which defaults to 500.
(2) \textit{scaling\_factor} ($\theta_{s}$): The scaling factor, which defaults to 0.5.
(3) \textit{min\_threshold and max\_threshold}: The minimum and maximum limits for the threshold value, which are 50 and 10,000, respectively.
(4) \textit{exponent} ($\beta$): The power exponent, is used to control the nonlinear scaling, which defaults to 0.5.
(5) \textit{scaling\_factor} ($\gamma$): The scaling factor.
(6) \textit{quantile\_threshold} ($\theta_{q}$): The quantile of video like number is set as 75\%.

The scaled threshold is calculated as
$$ \theta_{s} = \theta_{b} \times \left( \frac{N}{\theta_{q}} \right)^{\beta} \times \gamma,$$
where $N$ is the number of videos.
The \textit{Like\_threshold} ($\theta_{l}$) is calculated as
$$ \theta_{l} = \begin{cases} 
\text{min\_threshold}, & \text{if } \theta_{s} < \text{min\_threshold} \\
\text{max\_threshold}, & \text{if } \theta_{b} > \text{max\_threshold} \\
\theta_{s}, & \text{otherwise}
\end{cases}. $$

Videos with likes below the specified threshold are regarded as low-frequency. 
To mitigate the long tail issue, only those videos that surpass this threshold are preserved.

\subsection{Data Statistics and Quality}
\textbf{Data Statistics}. Table \ref{tab:data_analysis} provides statistics of {\dataset}. 
Statistics show that dialogues in KwaiChat are rich in dialogue numbers and long in video duration, which is much larger than other multi-modal dialogue datasets.

Figure~\ref{figure:category} illustrates the domains of {\dataset}.
The dataset is expansive, encompassing a diverse array of 30 domains, underscoring its richness and providing a robust foundation for the advancement of mixed-type dialogue systems.
There are six main domains, including lifestyle and interests, health and lifestyle, education and culture, entertainment and arts, technology and innovation, and news and current affairs.
There are a total of 30 domains in the dataset, 
with 5 domains for health and lifestyle,
6 domains for education and culture, 
9 domains for entertainment and arts, 
2 domains for technology and innovation,
3 domains for news and current affairs,
5 domains for lifestyle and interests.

\begin{table}[t]
\centering
\small
\begin{tabular}{@{}lcccccc@{}}
\toprule
Split                       & Language & \# of Videos & Avg-v       & Avg-u           \\ \midrule
\multirow{4}{*}{Train}      & POR       & 41,400        & 83.05        & 3.04                \\
                            & ZH       & 21,021        & 58.22        & 11.28               \\
                            & ID       & 11,419        & 57.33       & 3.00                 \\
                            & ES      & 3,824         & 35.12        & 3.14                  \\ \midrule 
\multirow{4}{*}{Validation} & POR       & 4,116         & 80.40        & 3.14                 \\

                            & ZH       & 2,117         & 58.43        & 11.28                  \\
                            & ID       & 1,073         & 59.25        & 2.97                   \\
                            & ES      & 371         & 31.90        & 3.00                     \\ \midrule
\multirow{4}{*}{Test}       & POR       & 4,209         & 84.24        & 3.66                  \\
                            & ZH       & 2,119         & 59.49        & 10.94                   \\
                            & ID       & 1,147         & 58.80        & 3.08                    \\
                            & ES      & 393         & 34.09        & 3.32                     \\ \bottomrule
\end{tabular}
\caption{Data division of KwaiChat. ``Avg-v'', and ``Avg-u'' denote the average duration of videos, the average utterance of dialogues. }
\label{tab:data_analysis}
\end{table}

\begin{figure*}[t]
	\small
	\centering
	\includegraphics[width=0.9\linewidth]{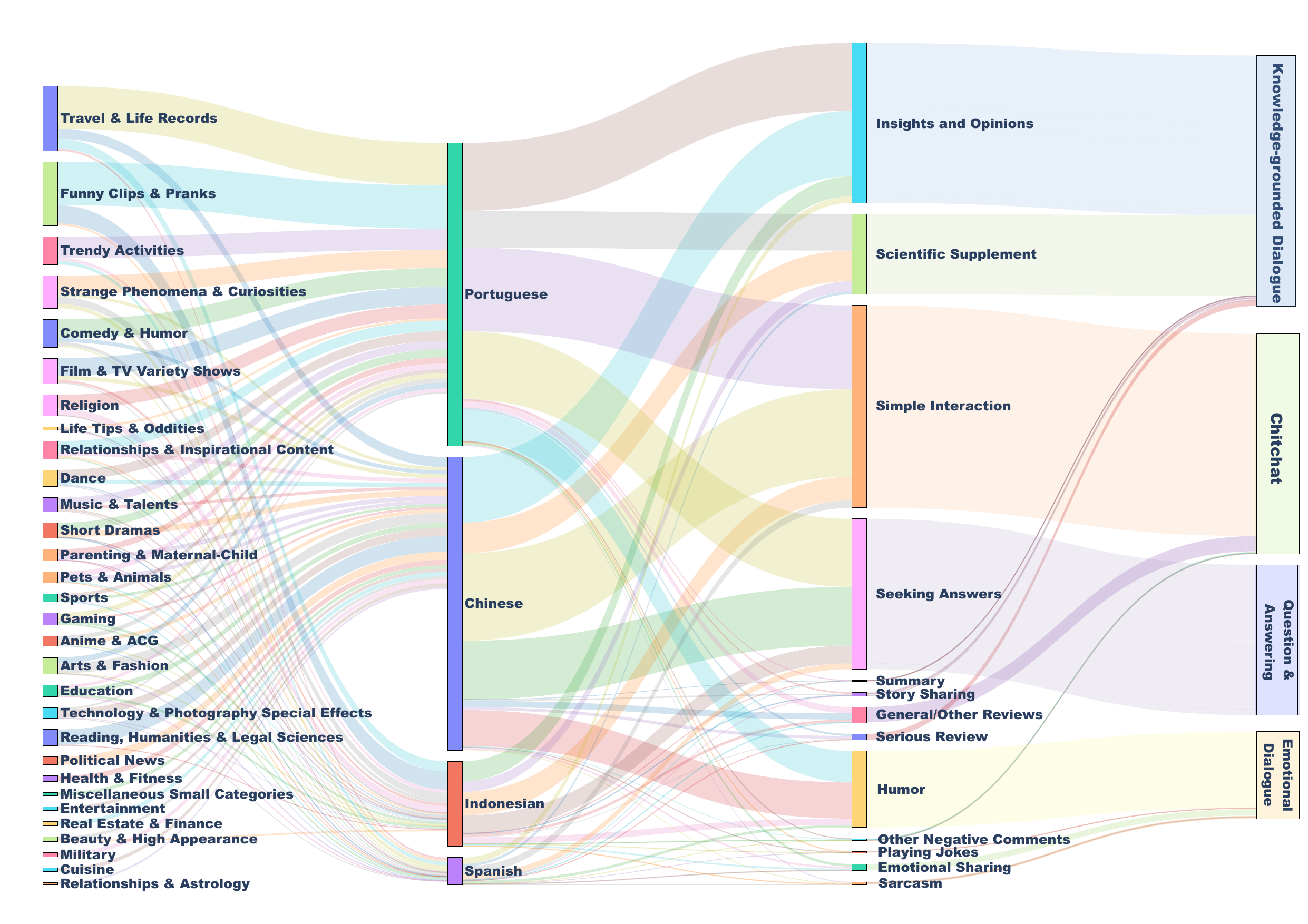}\\
	\caption{The domains, languages, topics, and dialogue types of {\dataset}. The first column lists the domains. The second column lists the four languages. The third column lists the topics. The fourth column lists the dialogue types.}
	\label{figure:domain-language-topic}
\end{figure*}

\begin{table*}[ht]
\centering
\small
\begin{tabular}{@{}llllllll@{}}
\toprule
                                    & Model                             & Language          & BLEU1 & BLEU2 & ROUGE-L & DIST-2 & DIST-3 \\ \midrule
\multirow{10}{*}{Text Model}                                    & \multirow{5}{*}{Qwen2-72B-Instruct} 
                                                                        & POR          & 2.25  & 0.21  & \textbf{5.00}    & 99.89  & 99.99  \\
                                    &                                   & ID           & 1.16  & 0.14  & 2.95    & \textbf{99.95}  & \textbf{100.00}  \\
                                    &                                   & ES           & 2.67  & 0.25  & 4.68    & 99.94  & 99.99  \\
                                    &                                   & ZH           & \textbf{5.85}  & \textbf{0.57}  & 0.23    & 99.62  & 99.90  \\
                                    &                                   & POR,ID,ES,ZH & 4.21  & 0.41  & 2.11    & 99.74  & 99.94  \\ \cmidrule(l){2-8} 
                                    & \multirow{5}{*}{LLaMA3.1-8B}  
                                                                        & POR          & 1.98  & 0.20  & \textbf{5.20}    & 99.10  & 99.30  \\
                                    &                                   & ID           & 0.94  & 0.14  & 1.77    & 95.86  & 94.84  \\
                                    &                                   & ES           & 2.67  & 0.30  & 4.12    & \textbf{99.61}  & 99.14  \\
                                    &                                   & ZH           & \textbf{5.16}  & \textbf{0.61}  & 0.16    & 99.49  & \textbf{99.57}  \\
                                    &                                   & POR,ID,ES,ZH & 3.72  & 0.43  & 2.02    & 99.05  & 99.06  \\ \midrule
\multirow{25}{*}{Multi-modal} & \multirow{5}{*}{Qwen-VL}            
                                                                        & POR          & 2.42  & 0.27  & \textbf{4.84}    & 90.84  & 92.40  \\
                                    &                                   & ID           & 0.93  & 0.14  & 2.23    & 79.31  & 81.78  \\
                                    &                                   & ES           & 2.00  & 0.28  & 3.65    & 76.29  & 78.59  \\
                                    &                                   & ZH           & \textbf{6.35}  & \textbf{0.93}  & 0.37    & \textbf{96.30}  & \textbf{97.51}  \\
                                    &                                   & POR,ID,ES,ZH & 4.56  & 0.64  & 1.99    & 92.58  & 94.04  \\ \cmidrule(l){2-8} 
                                    & \multirow{5}{*}{Video-LLaMA2-7B}     
                                                                        & POR          & 2.56  & 0.35  & \textbf{5.39}    & 77.97  & 80.21  \\
                                    &                                   & ID           & 1.49  & 0.22  & 3.32    & 79.24  & 82.38  \\
                                    &                                   & ES           & 2.29  & 0.27  & 4.49    & 80.15  & 82.32  \\
                                    &                                   & ZH           & \textbf{6.90}  & \textbf{1.11}  & 0.34    & \textbf{94.96}  & \textbf{96.20}  \\
                                    &                                   & POR,ID,ES,ZH & 4.92  & 0.77  & 2.31    & 87.82  & 89.56  \\ \cmidrule(l){2-8}
                                    & \multirow{5}{*}{InternVL2-26B}        
                                                                        & POR          & 1.86  & 0.23  & \textbf{4.04}    & 80.54  & 83.86  \\
                                    &                                   & ID           & 1.40  & 0.20  & 3.00    & 91.48  & 93.16  \\
                                    &                                   & ES           & 2.00  & 0.26  & 3.71    & 78.18  & 81.11  \\
                                    &                                   & ZH           & \textbf{6.30}  & \textbf{1.04}  & 0.45    & \textbf{94.82}  & \textbf{96.33}  \\
                                    &                                   & POR,ID,ES,ZH & 4.41  & 0.70  & 1.85    & 89.72  & 91.83  \\ \cmidrule(l){2-8} 
                                    & \multirow{5}{*}{LLaVA-Next-Video-7B} 
                                                                        & POR          & 0.48  & 0.05  & 1.18    & 36.80  & 38.16  \\
                                    &                                   & ID           & 0.49  & 0.07  & 1.22    & \textbf{66.13}  & \textbf{67.23}  \\
                                    &                                   & ES           & 0.77  & 0.08  & \textbf{1.63}    & 51.17  & 54.15  \\
                                    &                                   & ZH           & \textbf{1.98}  & \textbf{0.26}  & 0.30    & 62.60  & 63.30  \\
                                    &                                   & POR,ID,ES,ZH & 1.46  & 0.18  & 0.64    & 56.36  & 57.33  \\ \cmidrule(l){2-8} 
                                    & \multirow{5}{*}{GPT-4o}           
                                                                        & POR          & 2.68  & 0.27  & \textbf{5.71}    & 99.83  & \textbf{99.98}  \\
                                    &                                   & ID           & 1.23  & 0.16  & 3.05    & \textbf{99.87}  & 99.97  \\
                                    &                                   & ES           & 3.01  & 0.34  & 5.65    & 99.75  & 99.96  \\
                                    &                                   & ZH           & \textbf{6.34}  & \textbf{0.81}  & 0.26    & 99.52  & 99.79  \\
                                    &                                   & POR,ID,ES,ZH & 4.70  & 0.58  & 2.31    & 99.65  & 99.87  \\ \bottomrule
\end{tabular}
\caption{Results of 7 models 
(Qwen2-72B-Instruct~\cite{yang2024qwen2}, LLaMA3.1-8B~\cite{dubey2024llama}, Qwen-VL~\cite{bai2023qwen}, Video-LLaMA2-7B~\cite{zhang2023video}, InternVL2-26B~\cite{chen2024internvl}, LLaVA-Next-Video-7B~\cite{li2024llava}, and GPT-4o~\cite{openai2023gpt4}) 
on {\dataset} in the zero-shot setting. Results are reported in percentage (\%). ``POR'', ``ID'', ``ES'', and ``ZH'' stand for Portuguese, Indonesian, Spanish, and Chinese.}
\label{tab:results}
\end{table*}

Figure~\ref{figure:domain-language-topic} illustrates the domains, languages, topics, and dialogue types of~\dataset.
The table elegantly delineates the intricate connections between diverse elements,
which shows that our filtering algorithm ensures a balanced and uniform distribution across all categories.
Besides, the dataset contains 30 domains, underscoring its richness and providing a robust foundation for the advancement of mixed-type dialogue systems.
The topics are collected from Kwai, and each topic is categorized into one of four dialogue types. As shown in Figure~\ref{figure:sample}, dialogue types are automatically assigned based on the given dialogue topics. We conducted a manual review of the data, and the results showed high data quality.

\noindent
\textbf{Data Quality}. Following the evaluation method in previous work~\cite{liu2020towards}, we conduct human evaluations for data quality.\footnote{A dialog is rated with a score ``1'' if it is coherent, logical, grammatically correct, otherwise ``0''. Then, we ask three experts to judge the quality of 200 randomly sampled dialogs}
Finally, we obtain an average score of 0.90 on this evaluation set, indicating that the dataset has high quality.

\section{Experiments and Results}
This section introduces the experimental setting,
data and evaluation metrics, 
baselines,
automatic evaluations, human evaluations,
and the case study.

\begin{table*}[!ht]
\centering
\small
\begin{tabular}{@{}lcccccc@{}}
\toprule
\multicolumn{1}{c}{Model}               & Sensibleness & Specificity  & Vision-R  & Dialog-R  & Human-likeness  & Average   \\ \midrule
Human comment                           & \textbf{6.00}& 5.25& 4.71& \textbf{5.96}& \textbf{6.54}& \textbf{5.70}\\ \midrule
Qwen2-1.5B-Instruct                     & 1.55& 2.17& 1.29& 2.13& 2.54& 1.93\\ \midrule
Qwen2-7B-Instruct                       & 2.34& 3.38& 1.96& 3.46& 3.09& 2.84\\ \midrule
Qwen2-7B-Instruct-LoRA                  & 3.71& 3.04& 2.30& 4.00& 5.29& 3.67\\ \midrule
Qwen2-7B-Instruct-Full                  & 2.59& 1.79& 1.71& 2.54& 4.46& 2.62\\ \midrule
Qwen2-72B-Instruct           & 3.88& 4.54& 2.96& 4.54& 4.00& 3.98\\ \midrule
Qwen2-72B-Instruct-3-shot     & 3.05& 3.54& 2.63& 3.67& 3.46& 3.27\\ \midrule
Qwen2-72B-Instruct-4-shot     & 3.25& 3.96& 2.38& 4.50& 3.63& 3.54\\ \midrule
Qwen2-72B-Instruct-5-shot     & 3.25& 3.84& 3.00& 4.09& 3.46& 3.53\\ \midrule
LLaMA3.1-8B                             & 1.96& 2.21& 1.75& 2.38& 2.75& 2.21\\ \midrule
Qwen-VL-Chat                            & 2.75& 3.54& 3.08& 3.58& 2.50& 3.09\\ \midrule
VideoLLaMA2-7B                      & 2.67& 3.13& 2.38& 3.67& 3.75& 3.12\\ \midrule
InternVL2-8B                            & 3.63& 4.25& 4.46& 4.71& 3.58& 4.12\\ \midrule
InternVL2-8B-LoRA                       & 2.84& 2.17& 2.33& 3.00& 4.58& 2.98\\ \midrule
InternVL2-8B-Full                       & 4.50& 4.04& 3.63& 4.63& 5.75& 4.51\\ \midrule
InternVL2-26B                           & 4.29& 4.71& 4.21& 5.08& 4.42& 4.54\\ \midrule
LLaVA-Next-Video-7B                     & 1.17& 1.38& 1.21& 1.38& 1.25& 1.27\\ \midrule
GPT-4o                                  & 4.29& 4.25& 4.38& 5.00& 4.71& 4.53\\ \midrule
GPT-4o-3shot                            & 4.25& 4.42& 3.79& 5.21& 5.29& 4.59\\ \midrule
GPT-4o-4shot                            & 5.13& 5.04& 4.13& 5.84& 5.54& 5.13\\ \midrule
GPT-4o-5shot                            & 5.58& \textbf{5.38}& \textbf{4.79}& 5.71& 6.25& 5.54\\ \bottomrule
\end{tabular}
\caption{Human evaluation results on five aspects. The average scores are reported. ``Vision-R'' and ``Dialog-R'' stand for vision relevance and dialog relevance, respectively.}
\label{tab:human_evaluation}
\end{table*}

\subsection{Experimental Setting}
\noindent
\textbf{Experimental Setting}.
For Transformer, the implementation by HuggingFace~\footnote{https://github.com/huggingface/transformers} is utilized,
where the hyperparameters follow the default settings in the original Transformer~\citep{vaswani2017attention}.

\noindent
\textbf{Computing Platform}.
Our experiments are conducted on the workstation with an Intel Xeon E5 2.40 GHz CPU, 128 GB memory, an NVIDIA A100 GPU, and CentOS 7.2. 

\subsection{Data and Evaluation Metrics}
\subsubsection{Data} Following Tiktalk~\cite{lin2023tiktalk}, we split {\dataset} into the training set, the validation set, and the test set by randomly sampling data with the ratio of 10:1:1, and ensure that random and identical videos do not appear in different sets.

\subsubsection{Automatic Evaluation Metrics}
Following Tiktalk~\cite{lin2023tiktalk}, we adopt two types of automatic metrics to evaluate the performance of models from different perspectives, token similarity and diversity.

\noindent
\textbf{Similarity}. Three basic automatic evaluation metrics for generation tasks are utilized in this work, including BLUE1~\cite{papineni2002bleu}, BLUE2~\cite{papineni2002bleu}, and ROUGE-L~\citep{lin-2004-rouge}.
BLUE is the n-gram precision that is calculated as the ratio of the number of n-grams in the candidate translation that also appear in the reference translation to the total number of n-grams in the candidate translation.
ROUGE-L focuses on the longest common subsequence between the generated text and the reference text.
These metrics all measure the similarity between the generated responses and the ground truth via n-gram matching.

\noindent
\textbf{Diversity}.
Following previous work\cite{liu-etal-2021-du}, this paper adopts DIST-2 and DIST-3~\cite{li2015diversity} to measure the diversity of generated responses.
Specifically, DIST-2 focuses on bigrams, which are pairs of consecutive words, and calculates the ratio of unique bigrams in the generated text to the total number of bigrams.
Similarly, DIST-3 evaluates trigrams, which are sequences of three consecutive words. The metric calculates the ratio of unique trigrams to the total number of trigrams in the generated text.

\subsubsection{Human Evaluation Metrics}
Following~\citet{lin2023tiktalk}, five human evaluation metrics are utilized in this work: 

\noindent
\textbf{Sensibleness} measures whether the response is meaningful and coherent, and conforms to logic and common sense;

\noindent
\textbf{Specificity} measures whether the response is relevant to a specific context, provides sufficient details, and avoids being too vague or general;

\noindent
\textbf{Vision Relevance} examines the degree of correlation between the response and the multi-modal information in the video;

\noindent
\textbf{Dialog Relevance} measures fluency, relevancy, and logical consistency of each response when given the global context;

\noindent
\textbf{Human-likeness} examines the similarity between generated response and with corresponding human response from the perspectives of appropriateness, fluency, and proactivity.

Each metric has 7 levels, ranging from point 1 (worst) to point 7 (best). 
We randomly select the generation results of 100 dialogues from the test set and ask three annotators who are experts in this task to independently evaluate them. 
The final score is calculated as the mean of their scores.

\subsection{Baselines}
We carefully select a few strong baselines for comparison.
Specifically, two baselines for text-only dialogue generation (Qwen2-Instruct~\cite{yang2024qwen2}, LLaMA3.1~\cite{dubey2024llama})
and five baselines for multi-modal dialogue generation (Qwen-VL~\cite{bai2023qwen}, Video-LLaMA2~\cite{zhang2023video}, InternVL2~\cite{chen2024internvl}, LLaVA-Next-Video~\cite{li2024llava}, GPT-4o~\cite{gpt4}).

\subsection{Automatic Evaluation}
We conduct experiments in three settings, zero-shot, in-context, in-context learning, and fine-tuning.

\noindent
\textbf{Zero-shot}.
The results of zero-shot on automatic evaluation metrics are shown in Table~\ref{tab:results}.
Two text-based models (Qwen2-Instruct~\cite{yang2024qwen2}, LLaMA3.1~\cite{dubey2024llama}) and 5 multi-modal models (Qwen-VL~\cite{bai2023qwen}, Video-LLaMA2~\cite{zhang2023video}, InterVL2~\cite{chen2024internvl}, LLaVA-Next-Video~\cite{li2024llava}, GPT-4o~\cite{openai2023gpt4}) are included.
These LLMs directly utilize video, dialogue context, and zero-shot prompts as the input.
Results of BLUE1, BLUE2, ROUGE-L, DIST-2, and DIST-3 are reported in percentage.
The results show the following conclusions.
First, video-based LLMs perform better than text LLMs, which indicates that LLMs cannot make accurate responses without video information.
Second, LLMs perform better in Chinese than in other languages.
The reason is that during the pre-training phase, there is an abundance of high-quality Chinese training corpora, which allows LLMs to achieve better pre-training in Chinese.
Third, the performance of LLMs is high on DIST, and the reason is that LLMs can generate diverse and fluent responses.
Finally, we conducted additional tests on samples in Spanish and Portuguese and found that LLaVa performs better in Spanish. The reason may be that LLaVa has superior performance in Spanish.

\noindent
\textbf{In-context Learning}.
The results of LLMs' in-context learning performance are shown in Table~\ref{tab:fewshot_results}.
The results show the following conclusions.
First, showing more cases performs better.
The reason is that more cases help LLMs to better understand the questions and answers, thus LLMs can make better responses.
Second, GPT-4o performs better than Qwen2 even with the few-shot setting, 
which shows the strong power of GPT-4o. 

\noindent
\textbf{Fine-tuning}.
The results of LLMs' fine-tuning performance are shown in Table~\ref{tab:sft_results}.
The results show that fine-tuning with Lora performs better than fine-tuning full parameters.
The reason may be that fine-tuning all parameters may result in overfitting to the training data, leading to low generalization.

\subsection{Human Evaluation}
Table~\ref{tab:human_evaluation} shows the human evaluation results on the five evaluation metrics.
Three conclusions can be obtained from the results.
First, these models perform better in terms of human-likeness and dialogue relevance compared to the other three evaluation metrics.
This shows that LLMs are good at generating fluency and context-aware responses.
Second, multi-modal models outperform text-only models due to their enhanced ability to leverage video information, which results in superior dialogue outcomes.
Third, GPT-4o achieves the best performance, which shows that GPT-4o is currently the best LLM on this task.

\subsection{Case Study}
To further investigate the performance of the baselines, we present a response case in Table~\ref{tab:case}.
In the video, the girl demonstrates her proficiency in cooking, decorating, and gardening. 
The text model praises her for being familiar with various aspects, with semantic representations that fit well with the context and fluent expression. 
However, the response lacks specific content that she is particularly adept at. 
By contrast, multimodal models can perceive the content of the video, which allows them to include specific details in their responses, making them more contextually relevant and aligned with the video content.

Besides, results show that LLaMA3.1 refuses to reply to the second comment.
The possible reason is that LLaMA3.1 regards the comment as an illegal event.



\section{Conclusion}
This work first identified the issue of how to generate multi-participant video-driven mixed-type dialogue.
To address this issue, we introduced a novel task and built a corpus of human-to-human multi-participant video-driven multilingual mix-typed corpus, named {\dataset}.
To facilitate further research, we established baselines on {\dataset}.
Thorough experiments involving 7 popular LLMs on {\dataset} indicated that GPT-4o performed best, yet it still fell short in certain situations, even with the in-context learning and fine-tuning methods.
Next, we will promote cross-lingual research.

\section*{Acknowledgments}
Thanks for the insightful comments and feedback from the reviewers. This work was supported by the National Key R\&D Program of China (No. 2023YFF0725600), the National Natural Science Foundation of China (No. 62406015), and research funding from Kuaishou Technology.

\section*{Limitations}
This research is limited to an analysis of four major languages, focusing on LLMs' performance towards video-driven mixed-type multi-party conversations. 
Moving forward, we are eager to delve into the realm of languages in low resources. 
By studying LLMs' performance in minority languages, we aim to enhance LLMs' ability to handle a broader range of linguistic inputs, ultimately leading to more inclusive and sophisticated conversational AI systems.

\section*{Ethics Statement}
Regarding safety, we acknowledge the limitations of the current framework proposed and the potential risks associated with deploying them directly for users. 
We do not recommend the direct deployment of the fine-tuned models from this work into real-life situations; currently, they are only suitable for academic research. 
While we intend to develop models for the greater good of society, it is crucial to recognize that the dataset contains potentially problematic content, including toxic or biased material that could be used to generate negative or offensive content.

\bibliography{acl_latex}

\appendix


\section{In-context Learning Results}
The results of LLMs' in-context learning performance are shown in Table~\ref{tab:fewshot_results}.
The results show the following conclusions.
First, showing more cases performs better.
The reason is that more cases help LLMs to better understand the questions and answers.
Second, GPT-4o performs better than Qwen2 with the few-shot setting, 
which shows the strong performance of GPT-4o. 
Third, current models still fall short in terms of BLEU and ROUGE metrics, indicating that further improvement is necessary. 
This also underscores the significance of this dataset.

\section{Fine-tuning Results}
The results of LLMs' fine-tuning performance are shown in Table~\ref{tab:sft_results},
where the performance of LLMs fine-tuning with the Lora technique and fine-tuning full parameters are reported.
The results show that fine-tuning with Lora performs better than fine-tuning full parameters.
The reason may be that fine-tuning of all parameters may result in overfitting to the training data, leading to low generalization.
Besides, 

\begin{figure*}[ht]
	\small
	\centering
	\includegraphics[width=\linewidth]{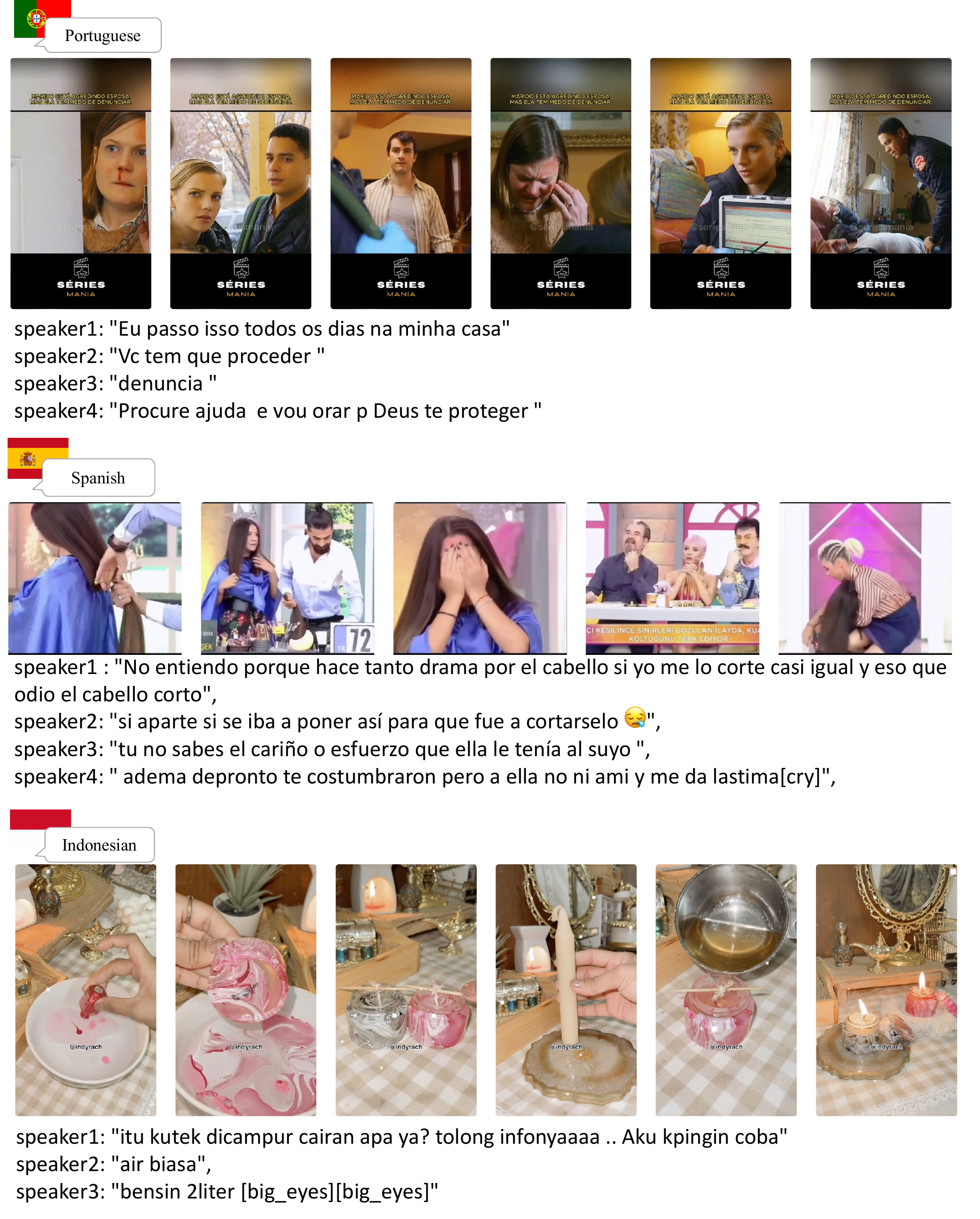}\\
	\caption{Examples of {\dataset}.}
	\label{figure:sample_other}
\end{figure*}

\begin{table*}[!ht]
\centering
\small
\begin{tabular}{@{}llllllll@{}}
\toprule
Model                                          & Setting                 & Language        & BLEU1 & BLEU2 & ROUGE-L & DIST-2 & DIST-3 \\ \midrule
\multirow{20}{*}{Qwen2-72B-Instruct-GPTQ-Int4} & \multirow{5}{*}{Base}   & POR             & 2.25  & 0.21  & \textbf{5.00}    & 99.89  & 99.99  \\
                                               &                         & ID              & 1.16  & 0.14  & 2.95    & \textbf{99.95}  & \textbf{100.0}  \\
                                               &                         & ES              & 2.67  & 0.25  & 4.68    & 99.94  & 99.99  \\
                                               &                         & ZH              & \textbf{5.85}  & \textbf{0.57}  & 0.23    & 99.62  & 99.90  \\
                                               &                         & POR,ID,ES,ZH    & 4.21  & 0.41  & 2.11    & 99.74  & 99.94  \\ \cmidrule(l){2-8} 
                                               & \multirow{5}{*}{3-shot} & POR             & 2.41  & 0.26  & \textbf{5.36}    & \textbf{99.93}  & 99.97  \\
                                               &                         & ID              & 1.31& 0.19& 3.35& 99.91& \textbf{99.99}\\
                                               &                         & ES              & 2.72& 0.27& 5.10& 99.89& \textbf{99.99}\\
                                               &                         & ZH              & \textbf{6.86}& \textbf{0.78}& 0.18& 99.11& 99.60\\
                                               &                         & POR,ID,ES,ZH    & 4.82& 0.55& 2.20& 99.46& 99.76\\ \cmidrule(l){2-8} 
                                               & \multirow{5}{*}{4-shot} & POR             & 2.31& 0.24& \textbf{5.29}& \textbf{99.94}& 99.99\\
                                               &                         & ID              & 1.35& 0.17& 3.40& 99.86& 99.96\\
                                               &                         & ES              & 2.44& 0.23& 4.65& 99.93& \textbf{100.0}\\
                                               &                         & ZH              & \textbf{6.97}& \textbf{0.83}& 0.29& 99.11& 99.59\\
                                               &                         & POR,ID,ES,ZH    & 4.87& 0.57& 2.27& 99.46& 99.76\\ \cmidrule(l){2-8} 
                                               & \multirow{5}{*}{5-shot} & POR             & 2.25& 0.25& \textbf{5.15}& \textbf{99.91}& \textbf{99.98}\\
                                               &                         & ID              & 1.35& 0.17& 3.43& 99.89& \textbf{99.98}\\
                                               &                         & ES              & 2.73& 0.29& 4.88& 99.84& 99.94\\
                                               &                         & ZH              & \textbf{7.06}& \textbf{0.83}& 0.17& 99.13& 99.61\\
                                               &                         & POR,ID,ES,ZH    & 4.91& 0.57& 2.17& 99.46& 99.77\\ \midrule
\multirow{20}{*}{GPT-4o}                       & \multirow{5}{*}{Base}   & POR             & 2.68  & 0.27  & \textbf{5.71}    & 99.83  & \textbf{99.98}  \\
                                               &                         & ID              & 1.23  & 0.16  & 3.05    & \textbf{99.87}  & 99.97  \\
                                               &                         & ES              & 3.01  & 0.34  & 5.65    & 99.75  & 99.96  \\
                                               &                         & ZH              & \textbf{6.34}  & \textbf{0.81}  & 0.26    & 99.52  & 99.79  \\
                                               &                         & POR, ID, ES, ZH & 4.70  & 0.58  & 2.31    & 99.65  & 99.87  \\ \cmidrule(l){2-8} 
                                               & \multirow{5}{*}{3-shot} & POR             & 2.72  & 0.38  & \textbf{6.55}    & \textbf{99.97}  & \textbf{99.89}  \\
                                               &                         & ID              & 1.73& 0.29& 4.33& 99.93& 99.76\\
                                               &                         & ES              & 2.74& 0.41& 5.95& 99.92& 99.81\\
                                               &                         & ZH              & \textbf{8.49}& \textbf{1.47}& 0.22& 98.30& 98.96\\
                                               &                         & POR, ID, ES, ZH & 5.95& 1.00& 2.68& 98.99& 99.29\\ \cmidrule(l){2-8} 
                                               & \multirow{5}{*}{4-shot} & POR             & 2.50& 0.35& \textbf{6.63}& \textbf{99.97}& 99.75\\
                                               &                         & ID              & 1.92& 0.28& 4.33& 99.95& 99.74\\
                                               &                         & ES              & 3.16& 0.57& \textbf{6.63}& \textbf{99.97}& \textbf{99.83}\\
                                               &                         & ZH              & \textbf{8.58}& \textbf{1.55}& 0.21& 98.07& 98.82\\
                                               &                         & POR, ID, ES, ZH & 6.01& 1.05& 2.69& 98.86& 99.21\\ \cmidrule(l){2-8} 
                                               & \multirow{5}{*}{5-shot} & POR             & 2.61& 0.41& \textbf{6.76}& \textbf{99.96}& 99.80\\
                                               &                         & ID              & 1.91& 0.28& 4.53& 99.93& 99.70\\
                                               &                         & ES              & 2.89& 0.47& 6.17& 99.95& \textbf{99.82}\\
                                               &                         & ZH              & \textbf{8.83}& \textbf{1.61}& 0.29& 97.96& 98.70\\
                                               &                         & POR, ID, ES, ZH & 6.18& 1.10& 2.78& 98.79& 99.15\\ \bottomrule
\end{tabular}
\caption{Results of Qwen2-72B-Instruct-GPTQ-Int4 and GPT-4o in the 3-shot, 4-shot, and 5-shot setting.}
\label{tab:fewshot_results}
\end{table*}

\begin{table*}[!ht]
\centering
\small
\begin{tabular}{@{}llllllll@{}}
\toprule
                                    & Model                                   & Language     & BLEU1 & BLEU2 & ROUGE-L & DIST-2 & DIST-3 \\ \midrule
\multirow{10}{*}{Text Model}        & \multirow{5}{*}{Qwen2-7B-Instruct-Lora} & POR          & 4.58  & 1.10  & \textbf{10.55}   & 86.17  & 71.37  \\
                                    &                                         & ID           & 3.85  & 0.89  & 7.70    & 88.07  & 72.82  \\
                                    &                                         & ES           & 5.58  & 1.33  & 10.21   & \textbf{94.01}  & 87.49  \\
                                    &                                         & ZH           & \textbf{9.93}  & \textbf{2.76}  & 0.27    & 92.44  & \textbf{93.27}  \\
                                    &                                         & POR,ID,ES,ZH & 7.58  & 2.03  & 4.47    & 90.13  & 84.40  \\ \cmidrule(l){2-8} 
                                    & \multirow{5}{*}{Qwen2-7B-Instruct-Full}  & POR         & 2.57  & 0.46  & \textbf{6.91}    & 90.75  & 69.99  \\
                                    &                                         & ID           & 1.77  & 0.33  & 3.96    & 82.32  & 63.64  \\
                                    &                                         & ES           & 3.65  & 0.52  & 5.82    & \textbf{94.10}  & 89.54  \\
                                    &                                         & ZH           & \textbf{9.39}  & \textbf{2.26}  & 0.05    & 91.93  & \textbf{92.99}  \\
                                    &                                         & POR,ID,ES,ZH & 6.40  & 1.47  & 2.73    & 90.79  & 83.07  \\ \midrule
\multirow{10}{*}{Multi-modal}       & \multirow{5}{*}{InternVL2-8B-LoRA}      & POR          & 3.23  & 0.51  & \textbf{7.49}    & 91.81  & 81.02  \\
                                    &                                         & ID           & 2.22  & 0.43  & 4.70    & 86.34  & 72.45  \\
                                    &                                         & ES           & 3.56  & 0.58  & 6.22    & \textbf{93.93}  & 86.17  \\
                                    &                                         & ZH           & \textbf{8.76}  & \textbf{2.02}  & 0.27    & 93.36  & \textbf{92.23}  \\
                                    &                                         & POR,ID,ES,ZH & 6.38  & 1.39  & 3.01    & 92.31  & 86.98  \\ \cmidrule(l){2-8}
                                    & \multirow{5}{*}{InternVL2-8B-Full}      & POR          & 3.17  & 0.60  & \textbf{7.75}    & 89.28  & 75.71  \\
                                    &                                         & ID           & 2.02  & 0.32  & 4.65    & 83.63  & 66.26  \\
                                    &                                         & ES           & 3.36  & 0.76  & 6.62    & 91.30  & 78.94  \\
                                    &                                         & ZH           & \textbf{8.29}  & \textbf{1.99}  & 0.31    & \textbf{93.56}  & \textbf{92.00}  \\
                                    &                                         & POR,ID,ES,ZH & 6.07  & 1.39  & 3.12    & 91.36  & 84.50  \\ \bottomrule
\end{tabular}
\caption{Results of Qwen2-7B and InternVL2-8B with Lora fine-tuning and full-parameter fine-tuning.}
\label{tab:sft_results}
\end{table*}

\section{Case Study}
To further investigate the performance of the baselines, we present a response case in Table~\ref{tab:case}.
Results show that LLaMA3.1 provides inconsistent answers to the main comment and refuses to reply to the second comment.
The possible reason is that LLaMA3.1 regards the second main comment as an illegal event.
QwenVL provides an inconsistent answer to the first main comment but provides a correct answer to the second one.
Video-LLaMA2 gives a wrong answer for both the two main comments, which shows that it contains serious hallucinations.
InternVL2 also contains hallucinations in the first comment.
Besides, only GPT-4o makes correct replies to these two comments.
The results show that current LLMs fall short in knowledge-based question-and-answering.

\begin{table*}[!ht]
\small
\centering
\begin{tabular}{@{}ll@{}}
\toprule
Main Comment &  \begin{CJK*}{UTF8}{gbsn}热知识：硝酸钾不与任何物质反应\end{CJK*} (Hot knowledge: Potassium nitrate does not react with any substance)                                                                                                                                                                                                                                                                                                                                      \\ \midrule
Ground Truth & \begin{CJK*}{UTF8}{gbsn}有硝酸根的大多不会\end{CJK*} (The majority of nitrate ions are not present)                                                                                                                                                                                                                                                                                                                                                                \\ \midrule
LLaMA3.1     & \begin{tabular}[c]{@{}l@{}} \begin{CJK*}{UTF8}{gbsn}很好奇哪些物质能和硝酸钾反应呢？\end{CJK*} \\ (I am curious about which substances can react with potassium nitrate?)  \end{tabular}                                                                                                                                                                                                                                                                                                                                \\ \midrule
QwenVL       & \begin{tabular}[c]{@{}l@{}}\begin{CJK*}{UTF8}{gbsn}硝酸钾可不，你试试把土豆丢进硝酸钾里，保证一个土豆变两个\end{CJK*}(Is potassium nitrate okay? \\You can try throwing the rice mud into potassium nitrate to make sure you can turn one into two)\end{tabular}                                                                                                                                                                                                                      \\ \midrule
Video-LLaMA2 & \begin{tabular}[c]{@{}l@{}}\begin{CJK*}{UTF8}{gbsn}硝酸钾确实不会与任何物质发生反应，这是一种非常稳定的化学物质。\end{CJK*}\\ (Potassium nitrate does not react with any substance, it is an extremely stable chemical substance.)\end{tabular}                                                                                                                                                                                                                                              \\ \midrule
InternVL2    & \begin{tabular}[c]{@{}l@{}}\begin{CJK*}{UTF8}{gbsn}哈哈，看来你对化学知识了解的挺多啊！不过，一种非常稳定的化学物质，它通常不会与大多数\end{CJK*}\\ \begin{CJK*}{UTF8}{gbsn}物质发生反应。但是，在高温下，它会与氢气反应生成氮氧化物和水。\end{CJK*} \\ (Haha, it seems that you have a good understanding of chemistry! However, potassium nitrate is indeed a \\ highly stable compound that typically does not react with most substances. However, at high temperatures, \\ it does react with hydrogen gas to produce nitrogen oxides and water.)\end{tabular} \\ \midrule
GPT-4o       & \begin{tabular}[c]{@{}l@{}}\begin{CJK*}{UTF8}{gbsn}哈哈，这个热知识简直太搞笑了，事实摆在眼前，硝酸钾可是剧烈反应的好手啊！\end{CJK*}(Haha, this hot \\ knowledge is simply hilarious! The fact is that potassium nitrate is a good hand for violent reactions!)\end{tabular}                                                                                                                                                                                                               \\ \midrule \midrule
Main Comment & \begin{CJK*}{UTF8}{gbsn}为什么加热前要把火焰摇几下？\end{CJK*}(Why do we need to lower the flame before heating?)                                                                                                                                                                                                                                                                                                                                                         \\ \midrule
Ground Truth & \begin{CJK*}{UTF8}{gbsn}防止试管受热不均匀爆炸\end{CJK*}(Prevent uneven heating and explosion of test tubes)                                                                                                                                                                                                                                                                                                                                                           \\ \midrule
LLaMA3.1     & \begin{CJK*}{UTF8}{gbsn}不好意思，无法为这个问题提供回复。\end{CJK*}(Sorry, we are unable to provide a response to this question.)                                                                                                                                                                                                                                                                                                                                           \\ \midrule
QwenVL       & \begin{tabular}[c]{@{}l@{}}\begin{CJK*}{UTF8}{gbsn}这一步骤是为了让硝酸甘油和硝酸钾混合物均匀受热，避免局部过热导致硝酸甘油分解产生气体。\end{CJK*} \\(This step is to evenly heat the mixture of nitric acid white oil and potassium nitrate, avoiding local  \\ overheating that may cause the decomposition of nitric acid white oil to produce gas.)\end{tabular}                                                                                                                         \\ \midrule
Video-LLaMA2 & \begin{tabular}[c]{@{}l@{}}\begin{CJK*}{UTF8}{gbsn}加热前把火焰摇几下可能是为了增加火焰的热量，从而加快加热的速度。\end{CJK*}  \\   (Shaking the flame before heating can increase the heat of the flame and accelerate the heating process.)\end{tabular}                                                                                                                                                                                      \\ \midrule
InternVL2    &  \begin{tabular}[c]{@{}l@{}} \begin{CJK*}{UTF8}{gbsn}因为这样可以让火焰更均匀，避免局部过热。\end{CJK*} \\ (Because this makes the flame uniform and avoid local overheating.) \end{tabular}                                                                                                                                                                                                                                                                                                                           \\ \midrule
GPT-4o       & \begin{tabular}[c]{@{}l@{}}\begin{CJK*}{UTF8}{gbsn} 嘿，看来你也是个化学爱好者！加热前摇几下火焰是为了让火焰更均匀，这样加热的时候不会\end{CJK*}\\ \begin{CJK*}{UTF8}{gbsn} 因为突然的温差导致试管破裂哦。科学实验真是处处有玄机呢！\end{CJK*}(Hey, it seems like you're also a \\ chemistry enthusiast! Shaking the flame before heating is to make the flame more stable and uniform, \\ so that the test tube will not rupture due to sudden temperature differences during heating. Scientific \\ experiments are full of opportunities!)\end{tabular}            \\ \bottomrule
\end{tabular}
\caption{Two generation cases of 5 LLMs for the given video and context.}
\label{tab:case}
\end{table*}

\end{document}